\documentclass[final,5p,times,twocolumn]{elsarticle}
\usepackage{amsmath,amsfonts}
\usepackage{algorithmic}
\usepackage{algorithm}
\usepackage{array}
\usepackage[caption=false,font=normalsize,labelfont=sf,textfont=sf]{subfig}
\usepackage{textcomp}
\usepackage{stfloats}
\usepackage{url}
\usepackage{verbatim}
\usepackage{graphicx}
\usepackage{enumitem}
\newlist{steps}{enumerate}{1}
\setlist[steps, 1]{label = Step \arabic*:}

\journal{}

\begin{document}

\begin{frontmatter}

\title{Motion-induced error reduction for high-speed dynamic digital fringe projection system}

\author{Sanghoon Jeon\corref{}}
\author{Hyo-Geon Lee\corref{}}
\author{Jae-Sung Lee\corref{}}
\author{Bo-Min Kang\corref{}}
\author{Byung-Wook Jeon\corref{}}
\author{Jun Young Yoon\corref{corresponding}}
\ead{junyoung.yoon@yonsei.ac.kr}
\author{Jae-Sang Hyun\corref{corresponding}}
\ead{hyun.jaesang@yonsei.ac.kr}

\cortext[corresponding]{Corresponding author}
\address{Department of Mechanical Engineering, Yonsei University, Seoul, 03722, South Korea}

\begin{abstract}
%Phase-shifting profilometry (PSP) is widely used and researched due to its simple setup, high accuracy, and high resolution. 
%However, any motion during the acquisition of fringe patterns can introduce errors because PSP assumes both the object and measurement system are stationary.
In phase-shifting profilometry (PSP), any motion during the acquisition of fringe patterns can introduce errors because it assumes both the object and measurement system are stationary. Therefore, we propose a method to pixel-wise reduce the errors when the measurement system is in motion due to a motorized linear stage.
The proposed method introduces motion-induced error reduction algorithm, which leverages the motor’s encoder and pinhole model of the camera and projector. 3D shape measurement is possible with only three fringe patterns by applying geometric constraints of the digital fringe projection system. We address the mismatch problem due to the motion-induced camera pixel disparities and reduce phase-shift errors. These processes are easy to implement and require low computational cost. Experimental results demonstrate that the presented method effectively reduces the errors even in non-uniform motion.
\end{abstract}


\end{frontmatter}


\section{Introduction}
Non-contact three-dimensional (3D) shape measurement is increasingly vital in various fields such as medical applications, 3D telecommunication, industrial detection, and robotics \cite{geng2011structured,newcombe2011kinectfusion,wang2019autonomous,zhang2021high}. Among the various 3D measurement methods, fringe projection profilometry (FPP), a structured light-based approach, is widely used and researched due to its simple setup, high accuracy, and high resolution \cite{wang2020review,marrugo2020state,xu2020status}. FPP has two main approaches: Fourier transform profilometry (FTP), which can utilize a single fringe pattern for 3D measurement, and phase-shifting profilometry (PSP), which uses multiple fringe patterns for more accurate 3D measurement \cite{zhang2018high,takeda1983fourier,creath1988phase}. While FTP is suitable for measuring high-speed phenomena, it is sensitive to noise and surface variations, leading to low measurement accuracy. Therefore, PSP, known for its robustness against noise and higher accuracy, is more commonly used.

However, it has limitations because phase-shifting algorithms assume both the object and the measurement system are stationary, and any motion during the acquisition of fringe patterns can introduce errors due to the assumption \cite{zhang2018highbook}. Accurately measuring dynamic scenes in 3D is a common and crucial requirement, especially in fields like mobile robotics \cite{kostavelis2015semantic}. Thus, many researchers have been conducting various studies to overcome the limitations \cite{lu2021motion}.

Firstly, the most intuitive approach is to capture fringe patterns at a speed significantly faster than the motion, reducing the motion-induced errors to a negligible level. Accordingly, researchers have conducted studies to enhance the acquisition speed of fringe patterns and improve the robustness of high-speed 3D measurement method \cite{gong2010ultrafast,heist2016high,zuo2018micro,wu2020high}. However, this approach has the drawback of increased hardware costs or low resolution, and it is not a fundamental solution, as the motion-induced errors can still occur if the motion exceeds the improved speed at which fringe patterns can be obtained.

Methods for reducing motion-induced errors under limited hardware conditions also have been researched. Some researchers proposed Fourier-assisted methods \cite{foucong2013accurate,foucong2014accurate,fouli2016motion,fouqian2019motion}.
Guo et al. \cite{fouguo2020real} proposed a method using a dual-frequency composite grating to detect the motion region. They then employed phase extraction through Fourier fringe analysis (FFA) to reduce motion-induced errors. This approach allows for improved 3D shape measurement by converting the PSP-based phase of distinguished motion regions to FFA-based phase, and is suitable for complex scenes with dynamic and static parts. However, it is not robust to changing motion states, and the FFA-based phase has inherent limitations of FTP. High-frequency details are smoothed, and accurately calculating low-frequency fringes is diffcult, resulting in lower measurement accuracy compared to PSP. Recently, there have been attempts in FPP to measure 3D shape using learning-based approaches with a single shot \cite{wang2021single,nguyen2022single,nguyen2023generalized}. However, challenges such as limited applicability to specific environments and relatively low accuracy still need to be addressed.

Lu et al. \cite{lu2013new} estimated the rotation matrix and translation vector representing motion by placing white circular markers around the object and tracking them. The estimated motion was used to correct the phase map. Later, to overcome the limitation of requiring markers, they proposed a method using the scale-invariant feature transform (SIFT) algorithm to estimate and reduce motion-induced errors \cite{lu2017automated}. However, both methods are limited to estimating only 2D motion, excluding the depth direction. Duan et al. \cite{duan2019phase} reduced errors using two reference planes and a moving marker, but this is also confined to 2D motion. Similarly, object tracking methods \cite{guo2018high,lu2020automated,duan2021automatic} are under investigation, and these methods are typically not effective in 3D non-uniform motion and often require high computational complexity.

Some researchers have proposed methods using iterative algorithms to predict motion and subsequently reduce the errors. Wang et al. \cite{wang2019motion} utilized the binary defocusing method and took advantage of additional temporal sampling by acquiring two fringe patterns per one projection cycle. They reduced phase error by using iterative algorithms and the difference between the original phase map and additional phase map.
Liu et al. \cite{one} estimated motion by comparing the positions of the same 3D points on an object in different frames. They effectively corrected pixel-wise phase-shift errors by utilizing the relationship between the derivative of the projector pin-hole model with respect to the world coordinate system and the estimated motion. However, this approach is limited to cases where the object's movement speed is not significantly faster than the camera's capture speed enough to neglect camera pixel disparities and the motion can be assumed to be uniform.

While previous methods have made efforts to estimate motion, alternative approaches have been proposed that do not involve estimating any parameters.
Wang et al. \cite{wang2018motion} discovered that the motion-induced phase error doubles the frequency of the projected fringe pattern. To address this, they employed the Hilbert transform to create another set of fringe patterns that were phase-shifted by $\pi/2$. By generating a new phase map from these Hilbert-transformed fringe patterns and averaging it with the original phase map, they successfully reduced phase errors. While this method has the advantage of not requiring explicit motion estimation, it has limitations as it does not consider the mismatch problem caused by camera pixel disparities and uses approximation method assuming the phase shift error is very small.
Recently, methods employing histogram equalization to reduce phase shift error have been proposed \cite{wang2022motion,wang2022efficient}. These methods leverage the characteristic that the period of the histogram of phase errors is $\pi$ in the phase domain. While these approaches offer the advantage of not requiring parameter estimation for the phase-shift errors, they do not consider the mismatch problem and assume a constant phase variation.

Many of the studies mentioned above have been conducted in scenarios where the measurement system is stationary, and the object is in motion. However, in contrast, there are cases where the measurement system, such as a robotic arm equipped with a 3D sensing sensor or a dental 3D scanner, is in motion. Therefore, we propose a method to pixel-wise reduce motion-induced errors in a phase-shifting profilometry system that is in motion due to a motorized linear stage. Our proposed method introduces a modified phase-shifting algorithm, considering motion-induced errors, and leverages the motor's encoder and pinhole model of the camera and projector. 3D shape measurement is possible with only three fringe patterns by applying geometric constraints of the digital fringe projection system. We address the mismatch problem by introducing a camera pixel correction process and resolve phase-shift errors, which are not assumed to be small, through a phase-shift error correction process. 
These processes are easy to implement and require low computational cost.
Experimental results demonstrated that our proposed method effectively reduces the errors even in non-uniform motion, providing undistorted texture maps and 3D measurement results.

\section{Principle}
In this section, we will introduce our proposed method in detail. First, the fundamental basis, the conventional phase-shifting algorithm will be described, and the sources of motion-induced error will be mathematically analyzed. Subsequently, we will present our method, which reduces the motion-induced error, in a step-by-step manner.

\subsection{Conventional phase-shifting algorithm}
The phase-shifting algorithm is widely used for accurate 3D measurement since it is robust to noise. For high-speed 3D shape measurement, the minimum number of fringe images, which is three, are often used since it reduces the error caused by the sequential movement of the system or object. The three fringe images of the conventional phase-shifting algorithm \cite{zhang2018highbook} can be described as,
\begin{align}
\label{conv_three}
&I_{1}(u^{c},v^{c})=I^{'}(u^{c},v^{c})+I^{''}(u^{c},v^{c})\cos[\phi(u^{c},v^{c})-\delta],\\
&I_{2}(u^{c},v^{c})=I^{'}(u^{c},v^{c})+I^{''}(u^{c},v^{c})\cos[\phi(u^{c},v^{c})],\\
&I_{3}(u^{c},v^{c})=I^{'}(u^{c},v^{c})+I^{''}(u^{c},v^{c})\cos[\phi(u^{c},v^{c})+\delta],
\end{align}where $(u^{c},v^{c})$ represents the location of the camera pixel, $I^{'}$ means the average intensity, $I^{''}$ means the modulation, $\delta$ is the phase shift and $\phi$ is the wrapped phase to be solved to locate the projector pixels. The conventional three-step phase-shifting algorithm uses the same location of the camera pixel $(u^{c},v^{c})$ for the three fringe images because it assumes that both the fringe projection system and the object being photographed remain stationary. However, this assumption is one of the sources of motion-induced error and introduces camera pixel error. The same holds true when using a constant phase shift which can introduce phase shift error.

If we use the phase shift $\delta=2\pi/3$ then the wrapped phase can be written as,
\begin{equation}
\label{conv_ph}
    \phi(u^{c},v^{c})=\tan^{-1}\left[\frac{\sqrt{3}(I_{1}-I_{3})}{2I_{2}-I_{1}-I_{3}}\right].
\end{equation}

The wrapped phase has a range $(-\pi,\pi]$ and $2\pi$ discontinuities by the four-quadrant inverse tangent function in (\ref{conv_ph}).
Therefore, to determine the continuous position of the projector pixels, it is necessary to obtain unwrapped phase map $\Phi(u^{c},v^{c})$ using unwrapping method such as temporal phase unwrapping or spatial phase unwrapping.
The purpose of the phase unwrapping method is to accurately locate the $2\pi$ discontinuities in the wrapped phase map and add the multiples of $2\pi$. It can be described as,
\begin{equation}
    \Phi(u^{c},v^{c})=\phi(u^{c},v^{c})+2\pi\times k(u^{c},v^{c}),
\end{equation}
where $k$ is an integer and commonly referred to as fringe order.

\subsection{Camera pixel error correction}
The positions of corresponding camera pixels in successive fringe images will vary if motion occurs during the acquisition of fringe patterns. This variation becomes more significantly as the object being photographed gets closer to the camera, introducing errors in phase-shifting profilometry. This problem is commonly referred to as mismatch problem between adjacent fringe images, and the error will be called camera pixel error.
Thus, in three-step phase-shifting algorithms, the positions of the camera pixels in preceding and subsequent fringe images should be corrected to match the position of the second fringe image's pixels along horizontal and vertical axis in the image plane because we use the second fringe image as a standard. The second fringe image is suitable as the standard since it contains the most overlap with both the first and third fringe images.
Therefore, instead of using $(u^{c},v^{c})$, the corrected camera pixels can be described as,
\begin{alignat}{2}
\label{camera_error1}
&u^{c}_{1}=u^{c}- \epsilon_{12}^{uc}(u^{c},v^{c}), \quad && v_{1}^{c}=v^{c}- \epsilon_{12}^{vc}(u^{c},v^{c}),\\
&u_{2}^{c}=u^{c}, \quad && v_{2}^{c}=v^{c}, \\
\label{camera_error3}
&u_{3}^{c}=u^{c}+ \epsilon_{32}^{uc}(u^{c},v^{c}), \quad && v_{3}^{c}=v^{c}+ \epsilon_{32}^{vc}(u^{c},v^{c}),
\end{alignat}
where $u_{3}^{c}$ means the corrected $u$-directional camera pixel of the third fringe image and $\epsilon_{32}^{uc}$ means the relative $u$-directional camera pixel error of the third fringe image with the second fringe image.
The sign of the camera pixel error term might be different depending on principle axis direction of the camera.
The pixels beyond the image range due to the camera pixel error term are discarded.
The camera pixel error term can be obtained using pin-hole model of a camera which can be described as,
\begin{multline}
\label{camera_pinhole}
s^{c}\begin{bmatrix}u^{c}\\v^{c}\\1\end{bmatrix}=\textbf{A}^{c}[\textbf{R}^{c},\textbf{t}^{c}]\begin{bmatrix}x^{w}\\y^{w}\\z^{w}\\1\end{bmatrix}\\=
\begin{bmatrix}C_{11}&C_{12}&C_{13}&C_{14}\\C_{21}&C_{22}&C_{23}&C_{24}\\C_{31}&C_{32}&C_{33}&C_{34}\\\end{bmatrix}\begin{bmatrix}x^{w}\\y^{w}\\z^{w}\\1\end{bmatrix},
\end{multline}
where $s^{c}$ is scaling factor, $\textbf{A}^{c}$ is intrinsic matrix of the camera and $\textbf{R}^{c}$ is rotation matrix and $\textbf{t}^{c}$ is translation vector between the camera coordinate system and world coordinate system. By utilizing camera coordinate system as the world coordinate system, we can set $\textbf{R}^{c}$ as the identity matrix and $\textbf{t}^{c}$ as the zero vector.

The camera pixel movement, referred to as the camera pixel error term, can be calculated through the differentiation of camera pin-hole model. If we have the $x^{w}$, $y^{w}$, and $z^{w}$ coordinates of a point in the world coordinate system, we can determine its position in the camera image coordinate system using (\ref{camera_pinhole}). Thus, we can describe how changes in the world coordinate values affect the pixel position in the camera image coordinate system through a set of partial differential equations as,
\begin{align}
\frac{\partial u^{c}}{\partial x^{w}}=
\frac{C_{11}(C_{32}y^{w}+C_{33}z^{w}+C_{34})-C_{31}(C_{12}y^{w}+C_{13}z^{w}+C_{14})}{(C_{31}x^{w}+C_{32}y^{w}+C_{33}z^{w}+C_{34})^{2}},
\end{align} where $x^{w}, y^{w}$, and $z^{w}$ are initially obtained using the conventional three-step phase-shifting algorithm. They can be used as the initial coarse reference. Subsequently, using the improved 3D results obtained after correction with our proposed method results in a more accurate outcome.

In other words, if we know the distance a point in the world coordinate system has moved, we can determine how much the camera pixel position corresponding to that point has changed, which is the camera pixel error, as follows:
\begin{align}
    \epsilon_{12}^{uc}(u^{c},v^{c})&=round\left[\frac{\partial u^{c}}{\partial x^{w}}\Delta x_{12}^{w}+\frac{\partial u^{c}}{\partial y^{w}}\Delta y_{12}^{w}+\frac{\partial u^{c}}{\partial z^{w}}\Delta z_{12}^{w}\right],\\
    \epsilon_{32}^{uc}(u^{c},v^{c})&=round\left[\frac{\partial u^{c}}{\partial x^{w}}\Delta x_{32}^{w}+\frac{\partial u^{c}}{\partial y^{w}}\Delta y_{32}^{w}+\frac{\partial u^{c}}{\partial z^{w}}\Delta z_{32}^{w}\right],
    %\epsilon_{12}^{vc}&=\frac{\partial v^{c}}{\partial x^{w}}\Delta x_{12}^{w}+\frac{\partial v^{c}}{\partial y^{w}}\Delta y_{12}^{w}+\frac{\partial v^{c}}{\partial z^{w}}\Delta z_{12}^{w}
\end{align}
where $round[]$ is the rounding operator used to make the value the nearest integer as the pixel's position is an integer, and $\Delta x_{12}^{w}$ means the moved distance between the third fringe image and the second fringe image on the x-axis of the world coordinates. It can be calculated as,
\begin{gather}
\label{change_vector}
    \vec{r} = [r_{x}, r_{y}, r_{z}]^\mathsf{T},\\
    \Delta x_{12}^{w} = r_{x}\times[d_{encoder}(t_{2})-d_{encoder}(t_{1})],
    %\Delta x_{12}^{w} &= r_{x}\int_{t_{1}}^{t_{2}}d_{encoder}(t)dt,
\end{gather}
where $\Vec{r}$ is one of the unit column vectors of the rotation matrix used to transform the motion system coordinates to the world coordinates, and it can be obtained by applying the plane fitting algorithm to the plane parallel to the motion direction. The number of such vectors needed may vary depending on the degrees of freedom in your motion system. The $d_{encoder}(t_{1})$ represents the traveled distance from the origin at time $t_{1}$ and can be calculated by synchronizing the motor's encoder with the camera external trigger.

Therefore, the camera pixel error due to the motion can be corrected, and the relationships as described in (\ref{relation1})-(\ref{relation3}) can be obtained after applying camera pixel correction.
\begin{alignat}{5}
\label{relation1}
&I_{1}^{'}(u_{1}^{c},v_{1}^{c})&& \approx  I_{2}^{'}(u^{c},v^{c}) && =I^{'}(u^{c},v^{c})&&\approx I_{3}^{'}(u_{3}^{c},v_{3}^{c})\\
\label{relation2}
&I_{1}^{''}(u_{1}^{c},v_{1}^{c})&& \approx I_{2}^{''}(u^{c},v^{c}) && =I^{''}(u^{c},v^{c})&&\approx I_{3}^{''}(u_{3}^{c},v_{3}^{c})\\
\label{relation3}
&\phi_{1}(u_{1}^{c},v_{1}^{c})&& \approx \phi_{2}(u^{c},v^{c}) && =\,\phi(u^{c},v^{c})&&\approx \phi_{3}(u_{3}^{c},v_{3}^{c})
\end{alignat}

In addition, if the camera pixel error and phase shift error is considered, then the general equations of the three fringe images in dynamic motion can be described as,
\begin{align}
\label{error_three1}
I_{1}(u_{1}^{c},v_{1}^{c})&=I_{1}^{'}(u_{1}^{c},v_{1}^{c})\nonumber\\
&+I_{1}^{''}(u_{1}^{c},v_{1}^{c})\cos[\phi_{1}(u_{1}^{c},v_{1}^{c})-\delta_{1}(u^{c},v^{c})],\\
\label{error_three2}
I_{2}(u^{c},v^{c})&=I_{2}^{'}(u^{c},v^{c})+I_{2}^{''}(u^{c},v^{c})\cos[\phi_{2}(u^{c},v^{c})],\\
\label{error_three3}
I_{3}(u_{3}^{c},v_{3}^{c})&=I_{3}^{'}(u_{3}^{c},v_{3}^{c})\nonumber\\
&+I_{3}^{''}(u_{3}^{c},v_{3}^{c})\cos[\phi_{3}(u_{3}^{c},v_{3}^{c})+\delta_{3}(u^{c},v^{c})],
\end{align}where $\delta_{3}(u^{c},v^{c})$ is the corrected phase shift of the third fringe image which means the motion-induced phase shift error is considered and phase shift is not a constant.

If we apply (\ref{relation1})-(\ref{relation3}) to (\ref{error_three1})-(\ref{error_three3}) respectively by using camera pixel correction, they can be simplified as,
\begin{align}
\label{sim_error_three1}
I_{1}(u_{1}^{c},v_{1}^{c})&=I^{'}(u^{c},v^{c})\nonumber\\
&+I^{''}(u^{c},v^{c})\cos[\phi(u^{c},v^{c})-\delta_{1}(u^{c},v^{c})],\\
\label{sim_error_three2}
I_{2}(u^{c},v^{c})&=I^{'}(u^{c},v^{c})+I^{''}(u^{c},v^{c})\cos[\phi(u^{c},v^{c})],\\
\label{sim_error_three3}
I_{3}(u_{3}^{c},v_{3}^{c})&=I^{'}(u^{c},v^{c})\nonumber\\
&+I^{''}(u^{c},v^{c})\cos[\phi(u^{c},v^{c})+\delta_{3}(u^{c},v^{c})].
\end{align}

\subsection{Phase shift error correction}
The constant phase shift value varies for each pixel due to the motion, and phase shift error occurs because (\ref{conv_ph}) is derived using the constant phase shift. Thus, it should be corrected by using phase shift error compensation terms.
The corrected phase shift can be described as,
\begin{align}
\label{ps_12}
    \delta_{1}(u^{c},v^{c})=\frac{2\pi}{3}+\epsilon_{12}^{up}(u^{c},v^{c}),\\
\label{ps_32}
    \delta_{3}(u^{c},v^{c})=\frac{2\pi}{3}+\epsilon_{32}^{up}(u^{c},v^{c}),
\end{align}
where $\epsilon_{32}^{up}$ is the phase shift error between the third fringe image and the second fringe image.

The phase shift error can be calculated in a similar way to the calculation of the camera pixel error, with the only difference being the use of the projector pin-hole model and phase units. Moreover, if the fringe patterns change only along $u^{p}$ direction on the projector, then it can be described as,
\begin{align}
    \epsilon_{12}^{up}(u^{c},v^{c})&=\frac{2\pi}{\lambda}(\frac{\partial u^{p}}{\partial x^{w}}\Delta x_{12}^{w}+\frac{\partial u^{p}}{\partial y^{w}}\Delta y_{12}^{w}+\frac{\partial u^{p}}{\partial z^{w}}\Delta z_{12}^{w}),\\
    \epsilon_{32}^{up}(u^{c},v^{c})&=\frac{2\pi}{\lambda}(\frac{\partial u^{p}}{\partial x^{w}}\Delta x_{32}^{w}+\frac{\partial u^{p}}{\partial y^{w}}\Delta y_{32}^{w}+\frac{\partial u^{p}}{\partial z^{w}}\Delta z_{32}^{w}),
\end{align} where $\lambda$ is the fringe pitch to be used to change pixel unit to phase unit.

In addition, if we apply (\ref{ps_12})-(\ref{ps_32}) equations to (\ref{sim_error_three1})-(\ref{sim_error_three3}) and simultaneously solving these equations, then the corrected wrapped phase map can be written as
\begin{align}
\label{general_meeq}
    \phi(u^{c},v^{c})=\tan^{-1}
    \left[\frac{A_{1}\cos(\delta_{1})+A_{2}\cos(\delta_{3})+A_{3}}{-A_{1}\sin(\delta_{1})+A_{2}\sin(\delta_{3})}\right]
\end{align}
with three intermediate variables as
\begin{align}
    &A_{1}=I_{3}(u_{3}^{c},v_{3}^{c})-I_{2}(u^{c},v^{c})\\
    &A_{2}=I_{2}(u^{c},v^{c})-I_{1}(u_{1}^{c},v_{1}^{c})\\
    &A_{3}=I_{1}(u_{1}^{c},v_{1}^{c})-I_{3}(u_{3}^{c},v_{3}^{c}).
\end{align}

If the motion is uniform or we can assume that the phase shift error $\epsilon_{12}^{up}(u^{c},v^{c})$ and $\epsilon_{32}^{up}(u^{c},v^{c})$ are same, then (\ref{general_meeq}) can be simplified as
\begin{align}
\label{uniform_meeq}
    \phi=\tan^{-1}\left[\frac{(2+\cos(\epsilon)+\sqrt{3}\sin(\epsilon))(I_{1}-I_{3})}{(\sqrt{3}\cos(\epsilon)-\sin(\epsilon))(2I_{2}-I_{1}-I_{3})}\right],
\end{align} where $\epsilon$ can be used as either $\epsilon_{12}^{up}(u^{c},v^{c})$ or $\epsilon_{32}^{up}(u^{c},v^{c})$.

\subsection{Phase unwrapping using geometric constraints}
\begin{figure}[!t]
\centering
\includegraphics[width=2.4in, height=2in]{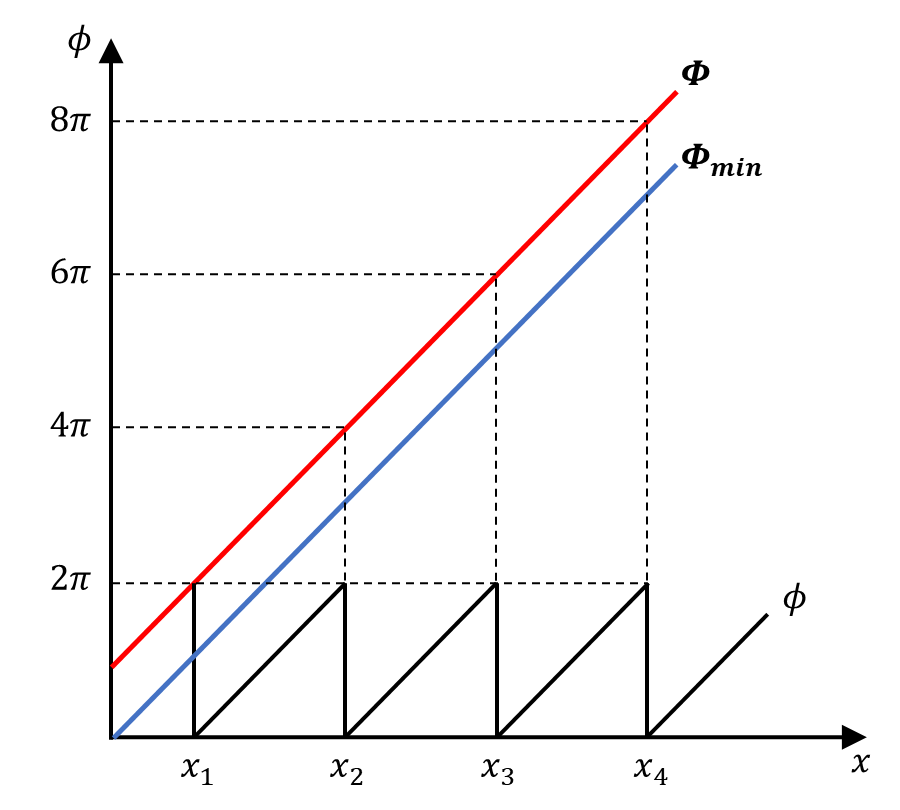}
\caption{Conceptual illustration of the method for removing $2\pi$ discontinuities in a wrapped phase map using the minimum phase map determined from geometric constraints \cite{an2016pixel}.}
\label{fig_zmin}
\end{figure}

\begin{figure}[!t]
\centering
\includegraphics[width=2.5in]{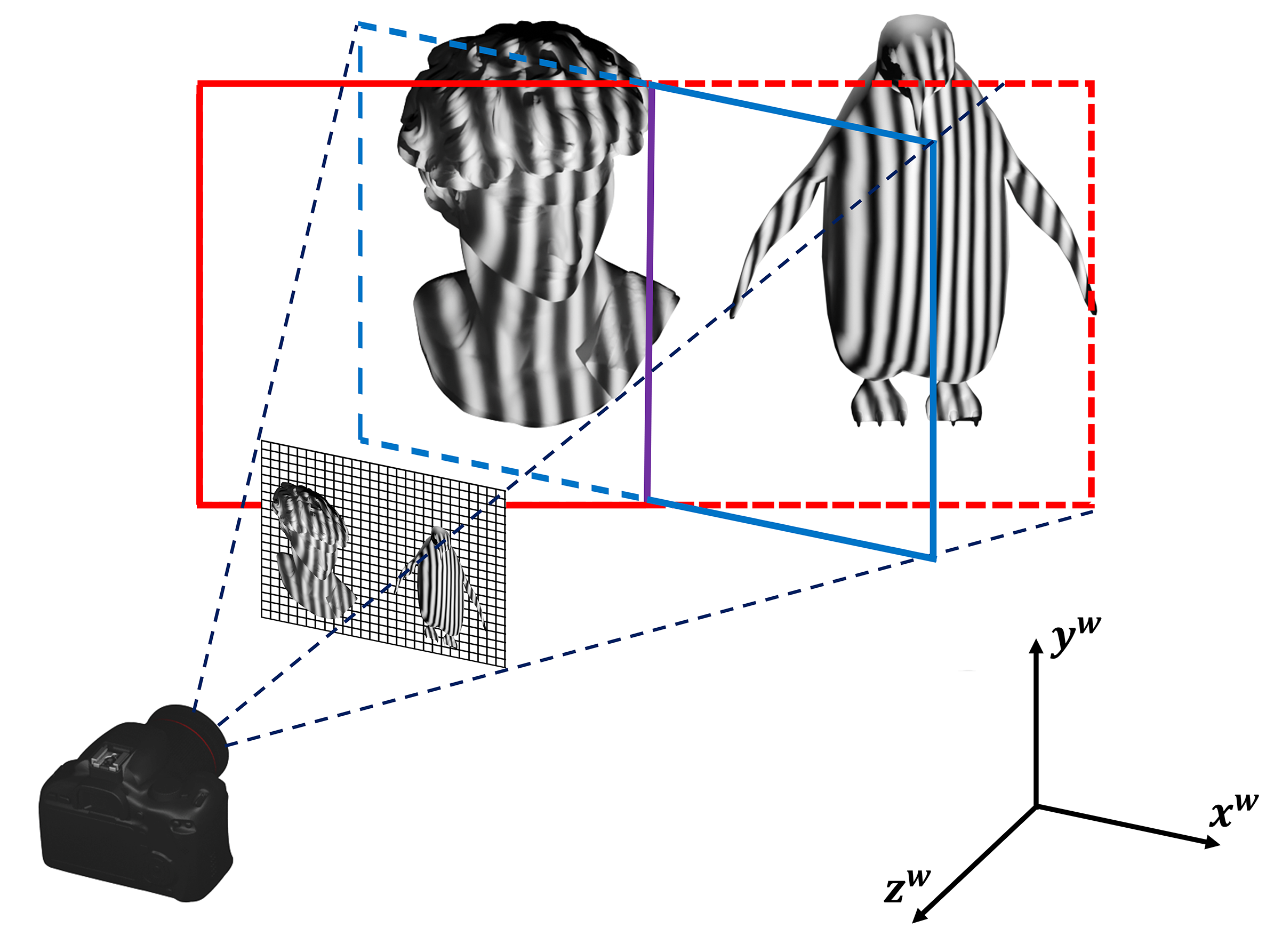}
\caption{Illustration of the rotation of the reference plane. The blue plane is the original reference plane, the red plane is the rotated reference plane and the purple line is the intersection line located at the center of the original reference plane.}
\label{fig_rotzmin}
\end{figure}

A temporal unwrapping method, proposed by An et al. \cite{an2016pixel}, leverages the geometric constraints of a structured light system to unwrap the wrapped phase without the need for additional image capture. This method utilizes the pinhole model of the camera and projector to establish geometric constraints. Specifically, using the camera's pinhole model, it generates a virtual plane at the closest depth of interest in world coordinate system, commonly referred to as $z_{min}$. For each camera pixel, the $x^{w}$, $y^{w}$, and $z^{w}$ values of this virtual plane can be obtained. Employing the projector's pinhole model, the absolute phase value corresponding to the each camera pixel, denoted primarily as $\Phi_{min}$, is determined. This value serves as the reference for unwrapping the wrapped phase. Given that the virtual plane is created based on the nearest distance in the area of interest, the unwrapping method can be applied under the assumption that the difference between the desired actual unwrapped phase $\Phi$ and $\Phi_{min}$ does not exceed $2\pi$. Furthermore, at discontinuous points, there is a difference of multiples of $2\pi$ or more between $\Phi_{min}$ and the wrapped phase value $\phi$, as shown in Fig. \ref{fig_zmin}. Therefore, fringe order $k$ can be determined as 
\begin{equation}
    k(u^{c},v^{c})=ceil\left[\frac{\Phi_{min} - \phi}{2\pi}\right],
\end{equation}where $ceil[]$ represents the ceiling operator, which rounds up to the nearest integer.

However, this method has a limitation in that it has a confined measurement depth range. Consequently, using a constant $z_{min}$ value can lead to errors in certain regions. For instance, when creating a virtual reference plane based on the left object, as shown in Fig. \ref{fig_rotzmin}, the distance between the right object and the virtual plane is too far for proper unwrapping. The same issue arises in the reverse scenario. Therefore, we employ a method that involves rotating the virtual reference plane at a specific angle.

First, the reference plane is created using the method proposed by An et al. \cite{an2016pixel}. Subsequently, we generate a rotation matrix based on the angle between the scanning object and the camera, which is then used to rotate the initial reference plane. If the scanning objects align parallel to the direction in which the structured light system moves, the vector in \text{(\ref{change_vector})} can be utilized. However, the rotation matrix was computed with the camera as the center of rotation. Two methods are employed to rotate the plane based on the center of the original reference plane: the first involves translating the rotated plane to align the centers of the rotated and original planes by adding the difference in z-axis distance between the centers of each plane. Subsequently, the depths of the rotated plane are used to create the final reference plane following the method  in \cite{an2016pixel}, as illustrated in Fig. \ref{fig_rotzmin}.

This method is highly effective when scanning stationary objects in a specific direction, as it allows using a single $z_{min}$ value across all frames, provided that the appropriate rotation angle is known.
Furthermore, the reference plane does not necessarily have to be in a plane shape. For example, in cases of rotational motion, it can be deformed and utilized in shapes like a semi-cylinder.

\subsection{Summary of method}
\begin{figure}[!ht]
\centering
\includegraphics[width=3in]{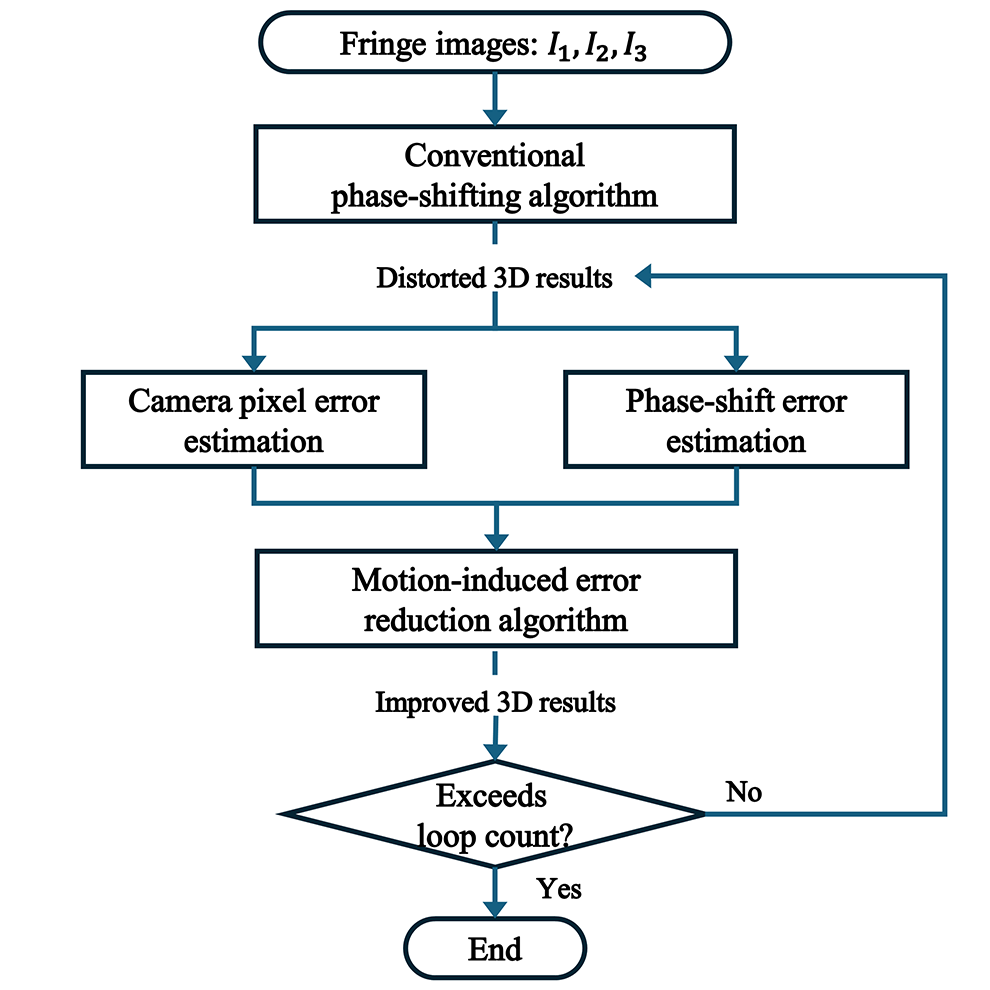}
\caption{Flowchart of our proposed method.}
\label{flowchart}
\end{figure}

To facilitate a visual understanding of our described approach, we illustrated a flowchart in Fig. \ref{flowchart}. 
The camera pixel error and phase-shift error are independently estimated, using the distorted 3D results obtained from the conventional method as coarse references. The estimated values for both errors are then utilized in the motion-induced error reduction algorithm, resulting in improved 3D outcomes.

In addition, the proposed method also can be summarized as follows:
\begin{enumerate}[label=Step \arabic*:]
    \item{Reconstruct 3D information using the conventional three-step phase-shifting method and phase unwrapping method using geometric constraints of structured light system.}
    \item{Perform camera pixel error correction, utilizing the values obtained in the previous step as initial values.}
    \item{Perform phase shift error correction.}
    \item{Improved results can be obtained after applying the phase unwrapping method in first step.}
    \item{Use the improved results as initial values and repeat steps 2 to 4 as desired.}
\end{enumerate}

\section{Experiments}
\begin{figure}[!t]
\centering
\includegraphics[width=3.4in, height=2.551in]{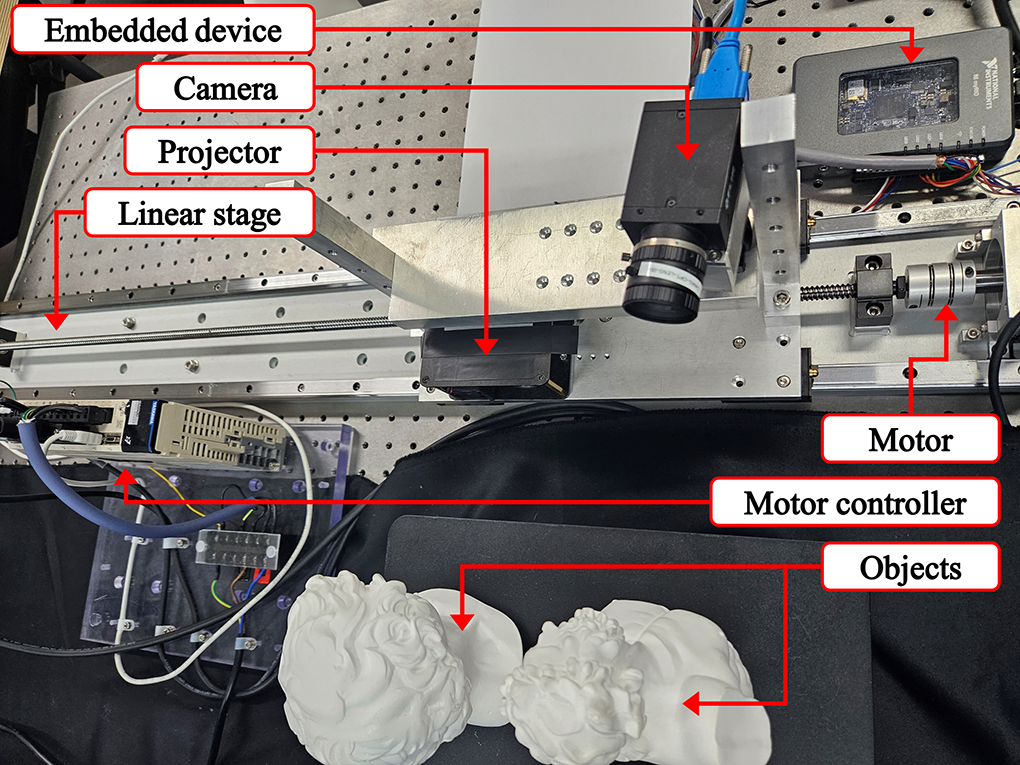}
\caption{Photograph of our experimental system setup.}
\label{setup_img}
\end{figure}
We built a phase-shifting profilometry (PSP) system and conducted several experiments to evaluate our proposed method. The system includes a complementary metal-oxide-semiconductor (CMOS) camera (model: FLIR Grasshopper3 GS3-U3-32S4M) that is attached with a 16 mm focal length lens (model: Computar M1614-MP2), and a digital light processing (DLP) projector (model: Texas Instruments LightCrafter 4500). The resolution of the projector was 912$\times$1140 pixels, while the camera's resolution was set at 1920$\times$1200 pixels. The system was calibrated using \cite{li2014novel}. The projector and the camera were synchronized at 120 Hz.

Furthermore, in order to simulate a dynamic environment for the PSP system, we integrated the system onto a linear stage as shown in Fig. \ref{setup_img}. This linear stage is driven by a servo motor (model: Yaskawa SGM7J-02AFD21) and controller (model: Yaskawa SGD7S-R70A00A), enabling movement in both the horizontal and depth directions of the camera coordinates. The camera and encoder of the motor were synchronized by using embedded device (model: National Instruments myRIO-1900).

\subsection{Uniform motion}
\begin{figure*}[!t]
\centering
\includegraphics[width=5.64in,height=5in]{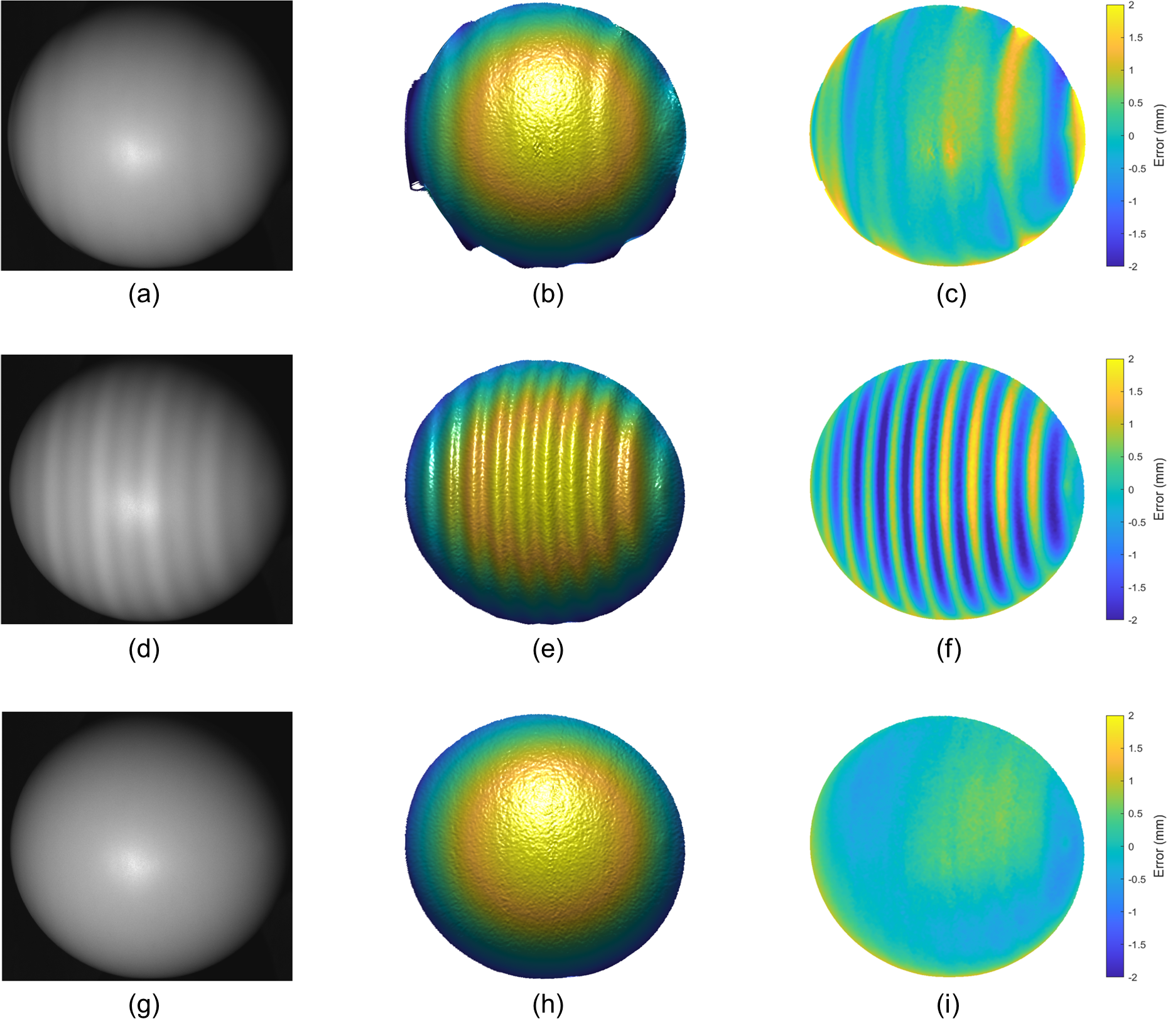}
\caption{Measurement results of a sphere while the PSP system is in uniform motion (associated with Visualization 1). (a) Texture map obtained using conventional phase-shifting method, (b) the corresponding 3D result, and (c) error map (mean: 0.176 mm, standard deviation: 0.537 mm, RMSE: 0.565 mm); (d) Texture map obtained only using camera pixel correction in our proposed method, (e) the corresponding 3D result, and (f) error map (mean: -0.313 mm, standard deviation: 0.987 mm, RMSE: 1.035 mm); (g) Texture map obtained using our proposed method, (h) the corresponding 3D result, and (i) error map (mean: 0.011 mm, standard deviation: 0.337 mm, RMSE: 0.337 mm).
}
\label{fig_sim}
\end{figure*}
To evaluate our proposed method, the PSP system moved at a constant speed of 80 mm/s, capturing images of a sphere with a diameter of 100 mm at a rate of 120 Hz. Subsequently, we compared the measurement results obtained using the conventional method with those from our proposed method. In this case, (\ref{uniform_meeq}) can be used to calculate the wrapped phase map because the motion is uniform. Finally, for a quantitative assessment, we compared the results with an ideal sphere as shown in Fig. \ref{fig_sim}. All 3D frames of the measurement results are displayed in Visualization 1.

To aid comprehension, we present some intermediate results of our proposed method in Fig. \ref{fig_sim}, displaying results in the order of conventional method result, camera pixel error correction result, and our proposed method, which includes both camera pixel error correction and phase-shift error correction.

Fig. \ref{fig_sim}(a) illustrates the texture map obtained using the conventional method. The texture of the sphere is distorted due to motion-induced measurement errors, leading to mismatches of the same pairs of camera pixels in adjacent fringe images. Thus, especially at the outer periphery of the sphere, significant pixel intensity variations between the mismatched object and background in adjacent images lead to distortions resembling motion blur. Correspondingly, Fig. \ref{fig_sim}(b) displays the distorted 3D result, and Fig. \ref{fig_sim}(c) presents the error map when compared to the ideal sphere. Prior to applying our method, the mean error is 0.176 mm with a standard deviation of 0.537 mm and root-mean-square error (RMSE) is 0.565 mm.

To address the mismatch problem, we employ our proposed camera pixel error correction process, eliminating distortions and achieving a spherical sphere in the texture map as shown in Fig. \ref{fig_sim}(d). However, vertical stripes caused by phase shift error become more pronounced than in Fig. \ref{fig_sim}(a). This phenomenon is attributed to the mismatched camera pixels, indicating that the camera pixel error partially offsets the phase shift error. Consequently, as the camera pixels in the fringe images are moved their correct positions, the only remaining phase shift error manifests as stripe-shaped distortions, visible in Fig. \ref{fig_sim}(e). The corresponding mean error is -0.313 mm with a standard deviation of 0.987 mm, and RMSE is 1.035 mm as shown in Fig. \ref{fig_sim}(f).

Finally, our proposed phase shift error correction process is applied to further reduce the remaining phase shift error. As shown in Fig. \ref{fig_sim}(g), the vertical stripes disappear, resulting in a perfectly shaped sphere in texture map. Fig. \ref{fig_sim}(h) shows a smooth 3D result, and when compared to the ideal sphere as shown in Fig. \ref{fig_sim}(i), the mean error is reduced to 0.011 mm with a standard deviation of 0.337 mm, indicating a significant reduction in error compared to the previous results.

\subsection{Non-uniform motion}

\begin{figure*}[!t]
\centering
\includegraphics[width=6.54in,height=3.4in]{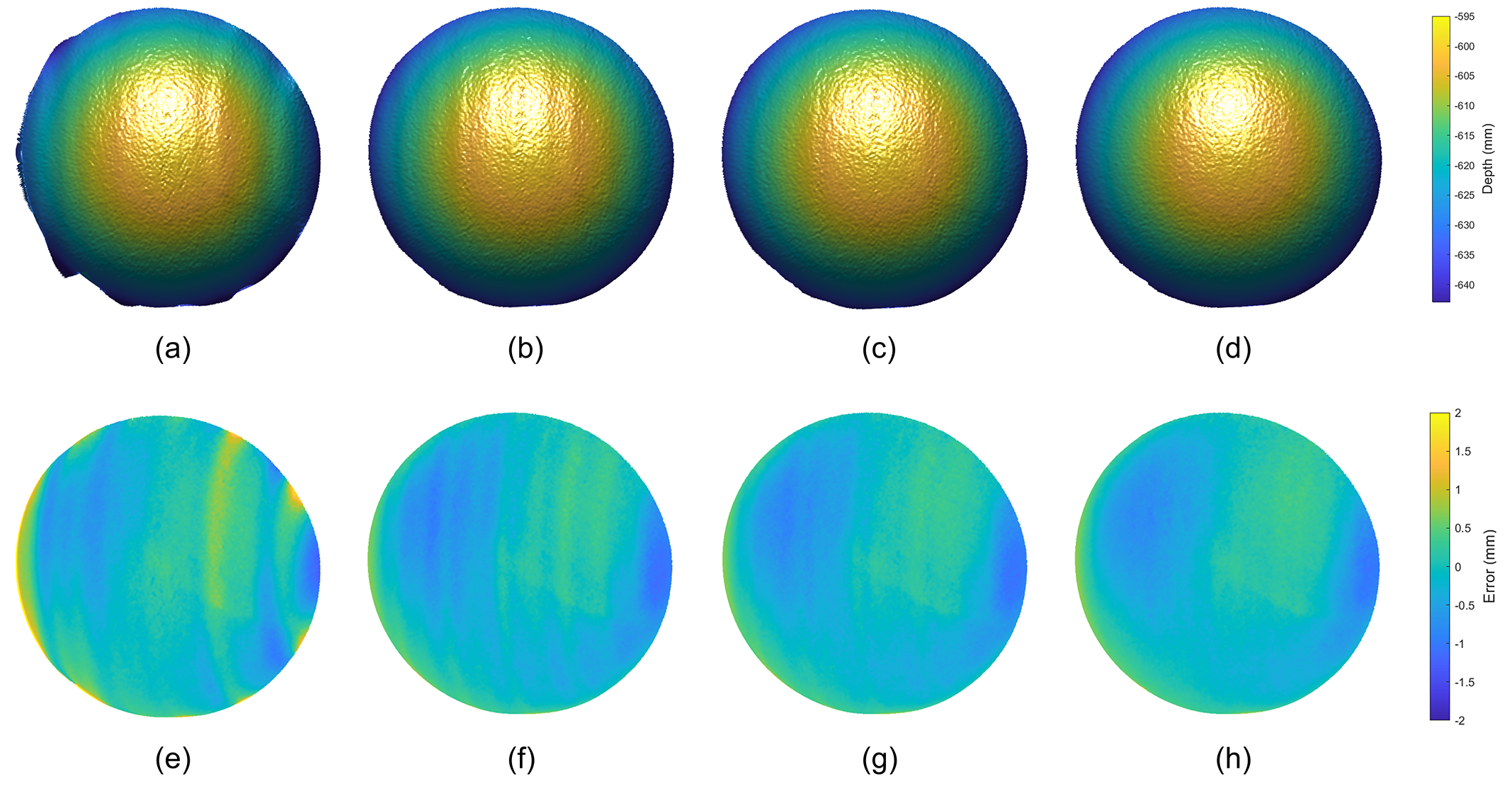}
\caption{Measurement results of a sphere while the PSP system is in non-uniform motion and in a stationary state (associated with Visualization 2). (a) 3D result from conventional phase-shifting method; (b) 3D result from proposed method for uniform motion; (c) 3D result from proposed method for non-uniform motion; (d) 3D result from another frame in a stationary state; (e) error map of the result in (a) (mean: -0.031 mm, standard deviation: 0.402 mm, RMSE: 0.403 mm); (f) error map of (b) (mean: -0.192 mm, standard deviation: 0.331 mm, RMSE: 0.382 mm); (g) error map of (c) (mean: -0.183 mm, standard deviation: 0.326 mm, RMSE: 0.374 mm); (h) error map of (d) (mean: -0.152 mm, standard deviation: 0.324 mm, RMSE: 0.358 mm).
}
\label{fig_sim2}
\end{figure*}

We conducted experiments to demonstrate the effectiveness of the proposed method not only in constant-speed intervals but also during acceleration intervals.
We used a sphere of the same specifications as the one used in the previous constant-speed experiment.
The PSP system underwent uniform acceleration, reaching 80 mm/s from a stationary state within 0.2 seconds, followed by constant-speed motion at 80 mm/s and then deceleration to a stop within 0.2 seconds. All 3D frames of the measurement results in these processes are displayed in Visualization 2. Among them, representative 3D frames are presented in Fig. \ref{fig_sim2}.

As seen in Fig. \ref{fig_sim2}(a), the 3D result obtained using the conventional phase-shifting method exhibits a distorted and deformed shape of the sphere. When compared to the ideal sphere, the mean error, as shown in Fig. \ref{fig_sim2}(e), is -0.031 mm with a standard deviation of 0.402 mm. Although the mean error appears to decrease compared to the mean error of the corrected results, this reduction is primarily influenced by outliers. Consequently, the standard deviation is relatively high and RMSE is 0.402 mm.

Fig.~\ref{fig_sim2}(b) presents the corrected 3D result using the same method as the previous constant-speed experiment, assuming uniform motion during acceleration. While there is some improvement compared to the results of the conventional method, noticeable distortions in the form of vertical stripe patterns still persist. This is attributed to assuming uniform motion during acceleration, which does not accurately reflect the situation as the movement speed exceeds the camera capture rate.
Therefore, in such cases, camera pixel errors ($\epsilon_{12}^{uc}$ and $\epsilon_{32}^{uc}$) as well as phase shift errors ($\epsilon_{12}^{up}$ and $\epsilon_{32}^{up}$) must be treated as distinct variables and calculated individually. 
Substituting each variable into (\ref{general_meeq}) yields the corrected wrapped phase map, resulting in the 3D result of our proposed method, as shown in Fig. \ref{fig_sim2}(c). Consequently, the improvements can be observed in terms of mean error, which decreased from -0.192 mm to -0.183 mm, standard deviation reduced from 0.331 mm to 0.326 mm, and RMSE improved from 0.382 mm to 0.374 mm, as shown in Fig. \ref{fig_sim2}(f) and Fig. \ref{fig_sim2}(g).

Fig. \ref{fig_sim2}(d) represents the 3D measurement results obtained when both the PSP system and the object are in a stationary state, while Fig. \ref{fig_sim2}(h) represents the error map corresponding to Fig. \ref{fig_sim2}(d). These results demonstrate similarity between the outcomes of our proposed method and those obtained in the stationary state.

\subsection{Objects with complex geometry}

\begin{figure*}[!t]
\centering
\subfloat[]{\includegraphics[width=3.2in, height=2in]{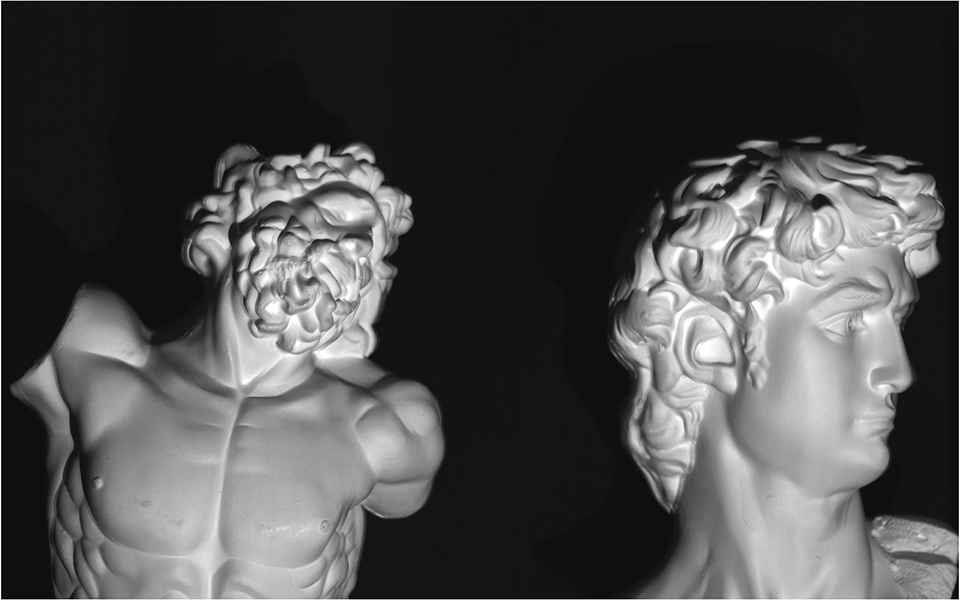}%
\label{fig_first_case3}}
\hfil
\subfloat[]{\includegraphics[width=3.2in, height=2in]{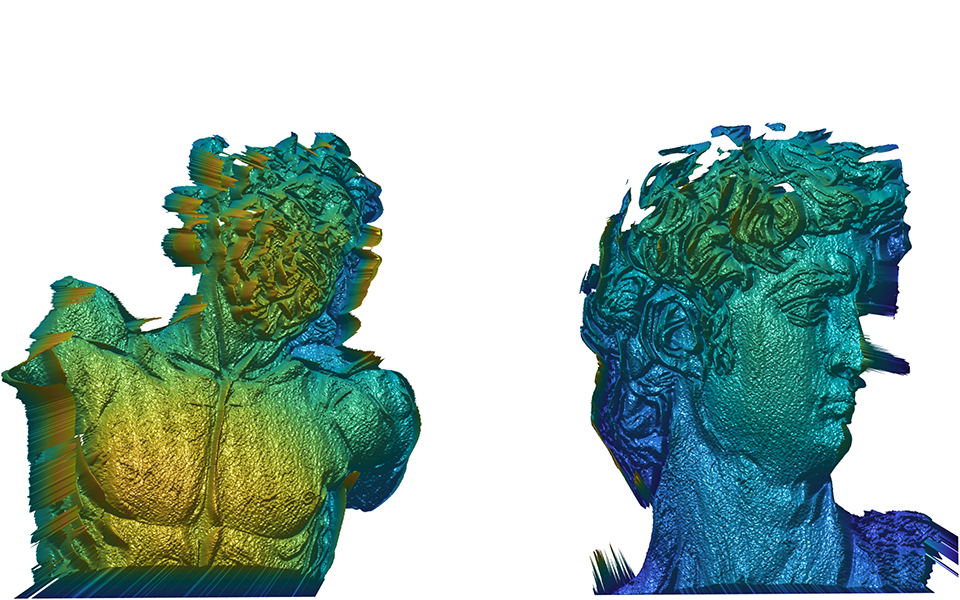}% bfig5
\label{fig_second_case3}}

\subfloat[]{\includegraphics[width=3.2in, height=2in]{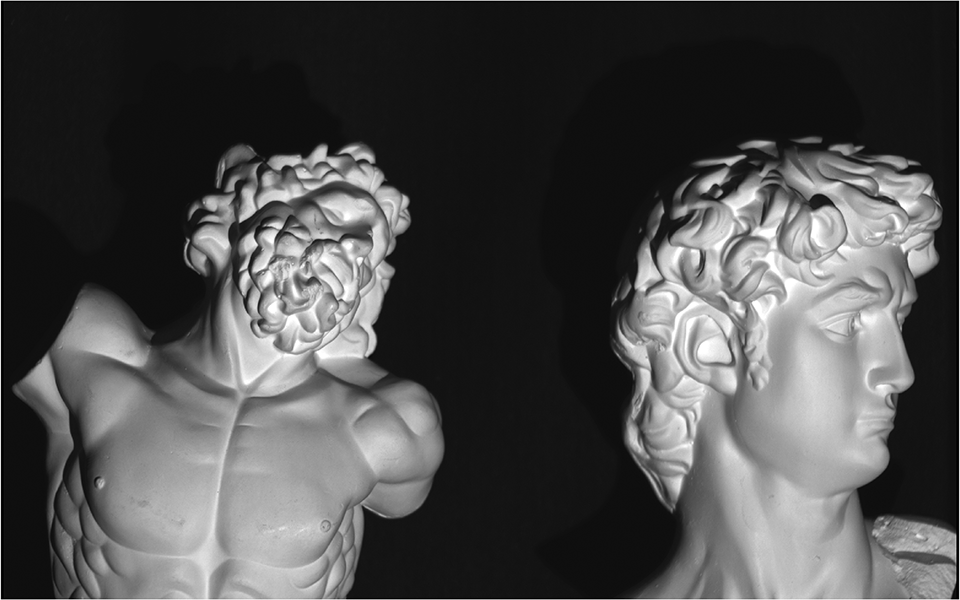}%
\label{fig_third_case3}}
\hfil
\subfloat[]{\includegraphics[width=3.2in, height=2in]{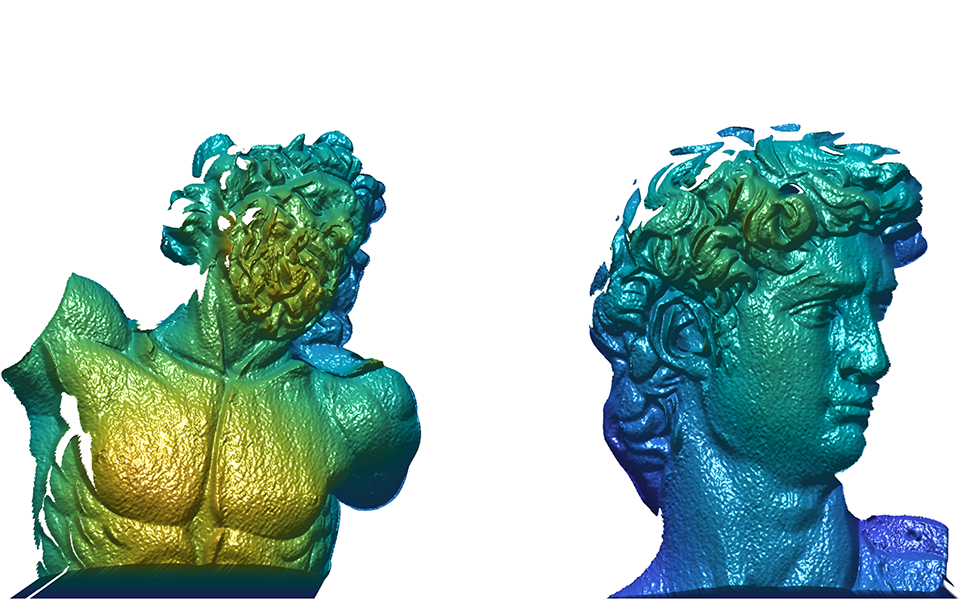}% 2
\label{fig_fourth_case3}}

\caption{Measurement results of objects with complex geometry while the PSP system is in uniform motion (associated with Visualization 3). (a) Texture map from conventional phase-shifting method and (b) the corresponding 3D result; (c) texture map from our proposed method and (d) the corresponding 3D result.
}
\label{fig_sim3}
\end{figure*}

Lastly, experiments were conducted to demonstrate the effectiveness of the proposed method in measuring complex shapes. The experiments were performed with the PSP system moving at a constant speed of 80 mm/s. The texture map and 3D result of a complex-shaped object obtained using the conventional method are presented in Fig. \ref{fig_sim3}(a) and Fig. \ref{fig_sim3}(b), respectively. Both the texture map and 3D result exhibit distortions and motion artifacts, especially noticeable in the facial area and boundary of the object. Upon closer inspection of the right object, distortions are evident in features such as the eyes, nose, mouth, and chin, appearing noticeably deformed compared to the actual shape. To mitigate these motion-induced errors, our proposed method was applied, and the resulting texture map and 3D results are shown in Fig. \ref{fig_sim3}(c) and Fig. \ref{fig_sim3}(d), respectively. Examining the texture map reveals the elimination of noise resembling motion blur observed in Fig. \ref{fig_sim3}(a), resulting in improved clarity. Additionally, a significant enhancement in the 3D result is evident.

All texture and 3D frames of the measurement results are displayed in Visualization 3. It's worth noting that shadow areas, where the pattern is not projected, may introduce artifacts; hence, certain shadow regions were masked and removed.

\section{Conclusion}
This article presented a novel method for reducing motion-induced error at pixel level in phase-shifting profilometry. The experimental results demonstrated that the proposed method efficiently reduces motion-induced error when the digital fringe projection system is in either uniform or non-uniform motion. It also works well even if the camera capturing speed is relatively slower than the motion because the method considers the mismatch problem between adjacent fringe image frames. Moreover, this method is suitable for real-time applications as it requires low computational cost and only three fringe images, which are the minimum number of patterns in phase-shifting profilometry.

We showed that our method can mitigate the motion-induced error as long as the motion between adjacent frames can be accurately estimated. Therefore, in the future, we plan to conduct research on estimating the motion between adjacent frames. This will demonstrate the applicability of our proposed method even in situations where both the digital fringe projection system and the scanning object are in motion.
\section*{Acknowledgements}
This research was supported by Yonsei University Research Fund of 2023-22-0434.
 \bibliographystyle{elsarticle-num} 
 \bibliography{ref}

\end{document}